# An Adaptive Behaviour-Based Strategy for SARs interacting with Older Adults with MCI during a Serious Game Scenario

ELEONORA ZEDDA, University of Pisa and HIIS Lab, ISTI-CNR, Italy

MARCO MANCA, HIIS Lab, ISTI-CNR, Italy

FABIO PATERNÒ, HIIS Lab, ISTI-CNR, Italy

CARMEN SANTORO, HIIS Lab, ISTI-CNR, Italy

The monotonous nature of repetitive cognitive training may cause losing interest in it and dropping out by older adults. This study introduces an adaptive technique that enables a Socially Assistive Robot (SAR) to select the most appropriate actions to maintain the engagement level of older adults while they play the serious game in cognitive training. The goal is to develop an adaptation strategy for changing the robot's behaviour that uses reinforcement learning to encourage the user to remain engaged. A reinforcement learning algorithm was implemented to determine the most effective adaptation strategy for the robot's actions, encompassing verbal and nonverbal interactions. The simulation results demonstrate that the learning algorithm achieved convergence and offers promising evidence to validate the strategy's effectiveness.

## 1 INTRODUCTION

With a senior population foreseen to more than double by 2050 worldwide [16], increasing demand for high-quality elderly support is likely to be expected in the coming years. In particular, Mild Cognitive Impairment (MCI) is an intermediate stage between the cognitive decline associated with normal ageing and more severe forms of dementia. Seniors with MCI often show memory loss or forgetfulness and may have issues with other cognitive functions such as language, attention and visual-spatial abilities. Currently, cognitive training for seniors with MCI is administered by professional caregivers who often use paper-based material. In this context, SAR can be a solution to cope with the problems of the tediousness of cognitive training and engage the user more during repetitive tasks. Socially assistive robots (SARs) assist human users through social interaction [6]. There has been increasing interest in using SARs in social contexts in recent years. Specifically, socially assistive robots obtained results improving the quality of life and engagement for the elderly and people with cognitive disabilities. In literature, various SARs have been used for engaging with older adults with cognitive impairments during cognitive training [11] [3] [2] [9]. Different studies [6] [7] aims to identify additional characteristics to implement social interactions, like expressing emotions, communicating with high-level dialogues, using natural cues, and performing distinctive personalities. Moreover, various studies [15] [13] found that a SAR with personalities can facilitate the interaction, as happens in human-human interaction during cognitive training by a human therapist. Emerging SARs may open up new possibilities in more effectively engaging Mild Cognitive impairment (MCI) older adults during repetitive cognitive training [9]. For these reasons, robot-performing personalities can be optimal for engaging the user more during HRI [17]. In this domain, some contributions indicate that personalised, tailored and adaptive robotic assistive systems can establish a productive interaction with the user, improving the effects of a therapy session. Some researchers [5] found that adaptive robot interactions are essential to provide comfortable and effective interactions with older adult users. An adaptive dialogue system would facilitate meaningful, effective communication and a more trusting relationship between the people with cognitive impairments and the robot [12] [14] [4] [10]. The previous studies designed adaptive strategies focusing mainly on exploring robotic dialogue strategies. In our study, we want to find the more suitable robot behaviour strategies composed of verbal and non-verbal parameters while exhibiting specific personalities on a SAR interacting with MCI older adults because, as the literature shows, adapting the SAR system can produce a more productive, engaging HRI, particularly for a sensitive target population like older adults with cognitive impairments. It is essential to provide a robot with an adaptive behaviour for this specific population because, as we can experience in previous studies [17] [9], we can see that elders and caregivers ask for a more natural and adaptive interaction with the robot during cognitive training.

## 2 ADAPTIVE ROBOT APPLICATION

Reinforcement learning is one of the most used methods to investigate the optimal behaviour for a SAR interacting with older adults [1]. The robot should always be able to show adequate behaviour according to the user state. Therefore, the robot should also be able to identify the user's state and adapt its response and behaviour to users with different cognitive capabilities. We designed a serious game scenario that stimulates specific areas affected by MCI. At the beginning of the session, the caregiver or psychologist will select one of the two personalities embedded in the robot. The personality will remain constant throughout the interaction. The use of serious game help to better engage the user, but a more natural and variant SAR behaviour can also create better interaction and environment for the user. For this reason, we designed and implemented a Q-learning algorithm and then implemented it in Pepper.

### 2.1 Key Elements in Scenario

**Robot Personality**. In our work, the humanoid robot used is the Pepper, developed by Softbank's Robotics. The robot exploits two personalities: an extraverted personality and an introverted personality. Studying the state of the art, we extrapolate different parameters that allow the robot to manifest such two opposite personalities, and we implement them on the Pepper[17]. Typically, extroverts tend to speak in a louder, faster, and higher-pitched manner. They are also more inclined to initiate conversations and talk more about themselves than others. Regarding body language, their gestures and movements are generally more expansive and faster and occur more frequently than those of introverted individuals. In the following scenario, there are three potential actions that the robot can take. If the user is engaged and responds correctly to the robot's query, the robot can react appropriately and enthusiastically to maintain the user's engagement. Alternatively, if the user appears disengaged, the robot should use stimulating behaviour to encourage the user to remain attentive and attempt to re-engage with the robot. The adaptation policy will dictate the appropriate course of action for the robot based on the user's current state during a serious game.

**Serious Game scenario**. The cooking game involves eight questions that challenge users to identify the correct sequence and weight of the ingredients. The serious game is divided into five stages: introduction, recipe instruction, question state, answer state and ending feedback. At the start of the application, the robot greets the user and asks if they are prepared to play. When the cooking game starts, the robot shows and vocally synthesises the ingredients for the selected recipe. The robot emphasises the sequential ingredients' order and weight during the recipe instruction. The quizzes follow, in which the user must use visual attention and working memory to identify the correct ingredients and choose them from the available options. The user interacts with the game using voice modality, and their state is assessed and evaluated after each of the eight questions. Based on the specific value obtained, the robot will take the optimal action for that state using the RL algorithm. We incorporated this algorithm because the robot's adaptability is vital to keep the user engaged throughout the serious game. In this context, an engaged robot adaptation is crucial to engage the user more during a serious game because, typically, the user is exposed to a series of repeated and standardised tasks and may create a high risk of dropping out of therapy and generating adverse conditions in the users, particularly older adults.

### 2.2 Modelling for reinforcement learning application

In the RL framework, interactions consist of a sequence of states (S), actions (A), and rewards (R). The RL agent observes the state $S_t$ from the environment and chooses an action $A_t$ to perform based on this observation. The chosen action is then executed, and the environment provides a new state, $S_{t+1}$, and a reward, $R_{t+1}$, to evaluate the transition. The agent's policy ($\pi$) maps states to actions and determines which action to select, given the current state. In our work, we use Q-learning to enable a robot to autonomously discover optimal behaviour through trial-and-error interactions with its environment in a real or simulated environment. It is a model-free, off-policy reinforcement learning that aims to find the best course of action given the agent's current state [8]. Q-Learning employs a Q table$(s, a)$ matrix to represent the policy, where $s$ corresponds to the current state of the environment, and $a$ represents the action selected by the robot to interact with the user. All available actions are defined within the robot's action space, denoted as A. In this specific scenario, an action $a$ belonging to $A$ a encompasses a combination of the robot's verbal feedback, vocal parameters, animations, and motor movements that contribute to its personality. According to the goals mentioned above, we define the key elements of the Markov decision process model as follows:

**State.** During a serious game scenario, a user's state is determined by its gaze direction (i.e. looking at 1:the robot, 2:tablet,2: up, 3:left, or 4:right); the smiling state (1:not smiling, 2:smiling, 3:broadly smiling), the given answer(right/wrong), and the level of engagement based on the previous parameters. In simulations, two user models were created (one for healthy users and one for users with MCI) to tailor the adaptation. The probability of giving the correct or wrong answer is also modelled based on the user models. To generate the next state, transition probabilities are defined in the algorithm based on the selected robot action and the user's engagement level.

**Action** The actions correspond to the robot's behaviour, which involves a combination of verbal feedback, vocal parameters, animations, and motor movements based on the supported robot's personality. During the scenario, the robot has three possible actions it can take. The first action (a0) involves generating dynamic and enthusiastic behaviour. For example, suppose the robot has an extroverted personality. In that case, it may provide more enthusiastic feedback, such as "Awesome! That's the correct answer! You're doing great!" with increased volume, speech rate, and pitch, as well as more extensive animations and more significant motor movements. The second action (a1) involves generating neutral robot behaviour. For example, suppose the robot has an extroverted personality. In that case, it may provide feedback such as "Good! That's the right answer!" with neutral vocal parameters defined by its personality and basic animations. The third action (a2) involves generating more stimulating behaviour. For instance, the robot may provide feedback such as "Correct answer! Let's keep up this level of attention!" with closer animations.

**Reward**. We have developed the immediate reward function following the purposes for a SAR in the cognitive game scenario: to stimulate the user to maintain a high level of engagement and to stimulate the user to concentrate on the questions. According to the simulated user model and the level of user engagement, the reward function considers the user's response to each question. Specifically, the robot should always try to prevent the user from getting trapped in a low level of engagement. The weights given to the algorithms are chosen to find the best policy to give to the application to maintain the user engagement level, at least in the medium or higher levels. This is important during repetitive cognitive training due to the tediousness of the tasks. Providing a more natural and engaging robot behaviour by putting the user's state in the centre and considering their cognitive state is essential to maintaining an engaged and participative user in this context. As said before, providing a more engaging robot behaviour for this particular population is critical because it can also improve the user's acceptability. The user can feel more comfortable focusing on the training and experience a more enjoyable interaction by interacting with a robot with some human behaviour or more natural actions.

### 2.3 Simulated Experiments

The q-table that the robot used to select the most appropriate action for that user state is presented in this section as our preliminary adaptation experiment results. This adaptation aims explicitly to keep user engagement high and to stimulate the user more if it dips to a low level. Based on the above explanation for the three essential components of Q learning (state, action, and reward function), we trained our reinforcement learning model in Python. It took the RL agent 100 epochs—each with 35 episodes—to complete its training. The discount factor and learning rate are set to 0.05 and 0.8, respectively. The *epsilon*-greedy policy was set to *epsilon* = 0.2. We started with an exponential *epsilon* decay. To penalize the algorithms at each step, we set a constant to -0.05. This is made to force the algorithms to discover the best policy quickly. The sum of Q-value updates for each epoch is used to assess the performance of our model, and the Q table means over the number of epochs is used to evaluate the convergence of performance. The simulation for both user models reaches the convergence of performance. Thus, reaching the convergence, additional training will not improve the model.

### 3 DISCUSSION

This work presents how we have designed and implemented a general RL algorithm to learn the adaptive verbal and non-verbal strategy for a SAR performing two personalities helping to cope in the cognitive training context of individuals with MCI. The model allows the robot's behaviour system to select an appropriate action given the user engagement level considering his cognitive state. We identify three possible actions per robot personality to stimulate and increase user engagement during a serious game scenario. For this reason, we have trained a Q table used in the intelligent algorithm to drive the robot by identifying the best action based on the detected user state. This work presents some limitations. The weight given to the algorithms was designed based on previous qualitative studies. From this perspective, in future work, we will closely collaborate with professional facilitators in this field and MCI users to adjust the reward function to ensure an effective, person-centred SAR policy using RL. The RL algorithm is implemented in the Pepper robot; however, to learn better the optimal policy, a real-world application with real users interacting with the SAR is necessary to evaluate the trained robot behaviour policy, which is our next step. Indeed, in the near future, we will conduct a trial in which the users will evaluate the adaptation system implemented in Pepper with MCI older adults in a within-subjects study. This study may allow the designing of a SAR behaviour adaptation policy to help older adults with MCI during a cognitive serious game application.

## REFERENCES
[1] Neziha Akalin and Amy Loutfi. 2021. Reinforcement learning approaches in social robotics. *Sensors* 21, 4 (2021), 1292.
[2] Linda M Beuscher, Jing Fan, Nilanjan Sarkar, Mary S Dietrich, Paul A Newhouse, Karen F Miller, and Lorraine C Mion. 2017. Socially assistive robots: measuring older adults' perceptions. *Journal of gerontological nursing* 43, 12 (2017), 35–43.
[3] Felix Carros, Johanna Meurer, Diana Löffler, David Unbehaun, Sarah Matthies, Inga Koch, Rainer Wieching, Dave Randall, Marc Hassenzahl, and Volker Wulf. 2020. Exploring human-robot interaction with the elderly: results from a ten-week case study in a care home. In *Proceedings of the 2020 CHI Conference on Human Factors in Computing Systems*. 1–12.
[4] Min Chi, Kurt VanLehn, Diane Litman, and Pamela Jordan. 2011. An evaluation of pedagogical tutorial tactics for a natural language tutoring system: A reinforcement learning approach. *International Journal of Artificial Intelligence in Education* 21, 1-2 (2011), 83–113.
[5] Dagoberto Cruz-Sandoval, Jesús Favela, Mario Parra, and Netzahualcoyotl Hernandez. 2018. Towards an adaptive conversational robot using biosignals. In *Proceedings of the 7th Mexican Conference on Human-Computer Interaction*. 1–6.
[6] David Feil-Seifer and Maja J Mataric. 2005. Defining socially assistive robotics. In *9th International Conference on Rehabilitation Robotics, 2005. ICORR 2005*. IEEE, 465–468.
[7] Terrence Fong, Illah Nourbakhsh, and Kerstin Dautenhahn. 2003. A survey of socially interactive robots. *Robotics and autonomous systems* 42, 3-4 (2003), 143–166.
[8] Jens Kober, J Andrew Bagnell, and Jan Peters. 2013. Reinforcement learning in robotics: A survey. *The International Journal of Robotics Research* 32, 11 (2013), 1238–1274.
[9] Marco Manca, Fabio Paternò, Carmen Santoro, Eleonora Zedda, Chiara Braschi, Roberta Franco, and Alessandro Sale. 2021. The impact of serious games with humanoid robots on mild cognitive impairment older adults. *International Journal of Human-Computer Studies* 145 (2021), 102509.
[10] Hamidreza Modares, Isura Ranatunga, Frank L Lewis, and Dan O Popa. 2015. Optimized assistive human–robot interaction using reinforcement learning. *IEEE transactions on cybernetics* 46, 3 (2015), 655–667.
[11] Olimpia Pino, Giuseppe Palestra, Rosalinda Trevino, and Berardina De Carolis. 2020. The humanoid robot NAO as trainer in a memory program for elderly people with mild cognitive impairment. *International Journal of Social Robotics* 12 (2020), 21–33.
[12] Chloé Pou-Prom, Stefania Raimondo, and Frank Rudzicz. 2020. A conversational robot for older adults with alzheimer's Disease. *ACM Transactions on Human-Robot Interaction (THRI)* 9, 3 (2020), 1–25.
[13] Adriana Tapus, Cristian Țăpuș, and Maja J Matarić. 2008. User—robot personality matching and assistive robot behavior adaptation for post-stroke rehabilitation therapy. *Intelligent Service Robotics* 1 (2008), 169–183.
[14] Konstantinos Tsiakas, Maria Dagioglou, Vangelis Karkaletsis, and Fillia Makedon. 2016. Adaptive robot assisted therapy using interactive reinforcement learning. In *Social Robotics: 8th International Conference, ICSR 2016, Kansas City, MO, USA, November 1-3, 2016 Proceedings 8*. Springer, 11–21.
[15] Sarah Woods, Kerstin Dautenhahn, Christina Kaouri, René te Boekhorst, Kheng Lee Koay, and Michael L Walters. 2007. Are robots like people?: Relationships between participant and robot personality traits in human–robot interaction studies. *Interaction Studies* 8, 2 (2007), 281–305.
[16] World Health Organization. 2019. Dementia. *World Health Organization* (2019). https://www.who.int/news-room/fact-sheets/detail/dementia
[17] Eleonora Zedda, Marco Manca, and Fabio Paternò. 2021. A Cooking Game for Cognitive Training of Older Adults Interacting with a Humanoid Robot.. In *CHIRA*. 271–282.